\documentclass[journal]{IEEEtai}

\usepackage{color,array}

\usepackage{amsmath,amsfonts}
\usepackage{amssymb}
\usepackage{textcomp}
\usepackage{booktabs}
\usepackage{stfloats}
\usepackage{url}
\usepackage{verbatim}
\usepackage{graphicx}
\usepackage{multirow}
\usepackage{tabularx}
\usepackage{cite}
\usepackage[colorlinks,urlcolor=blue,linkcolor=blue,citecolor=blue]{hyperref}

\begin{document}

\title{Instance-Aware Algorithm Selection for Maximum Clique via a Dual-Channel Graph Neural Architecture}

\author{
    Xiang Li, Shanshan Wang, and Chenglong Xiao\IEEEauthorrefmark{1}\\
    College of Mathematics and Computer, Shantou University, Shantou, China
    \thanks{\IEEEauthorrefmark{1}Corresponding author: Chenglong Xiao (email: chlxiao@stu.edu.cn).}
}

\markboth{}
{}

\maketitle

\begin{abstract}
Although the Maximum Clique Problem (MCP) has been extensively studied and features a rich ecosystem of exact solvers, empirical evidence shows that solver performance varies substantially across graph families. Consequently, selecting an appropriate algorithm for a given instance remains an open and practically important challenge that has received little systematic attention. We address this gap by developing an instance-aware selection framework that systematically combines global statistical descriptors with learned topological representations. We construct a comprehensive benchmark by evaluating four state-of-the-art exact solvers on a diverse collection of graph instances and deriving both global statistical and local structural features. An evaluation of conventional classifiers establishes Random Forest as a strong baseline and reveals that connectivity and topological features are key predictors of performance. Motivated by these observations, we introduce a dual-channel architecture that jointly leverages a Graph Attention Network for capturing local neighborhood patterns and a Multilayer Perceptron for modeling global statistical features. Extensive experiments show that the proposed dual-channel model consistently surpasses classical baselines and the single-best solver, achieving 90.43\% test accuracy. These findings demonstrate the value of integrating local topological encoding with global statistical cues for combinatorial algorithm selection. Code and models are available at: \url{https://anonymous.4open.science/r/GAT-MLP-7E5F}.
\end{abstract}

\begin{IEEEImpStatement}
The Maximum Clique Problem (MCP) is a computational task that appears in many real-world applications: finding fraud rings in financial networks, detecting coordinated bot campaigns on social media, and identifying protein interaction clusters in drug discovery. This paper introduces a system that learns to pick the right solver automatically, based on the structure of the input data. In experiments, the system selects the best solver with 90.43\% accuracy and runs substantially faster than relying on any single default solver. For difficult instances where a typical solver would time out, our selector identifies a more suitable algorithm and solves the instance substantially faster—often reducing hours of computation to minutes. More broadly, the approach offers a template for automating solver selection in other computationally hard problems where no one-size-fits-all solution exists.
\end{IEEEImpStatement}


\section{Introduction}
\label{sec:intro}

\IEEEPARstart{T}{he} Maximum Clique Problem (MCP) is a canonical NP-hard problem in graph theory with broad theoretical significance and real-world applications in areas such as bioinformatics \cite{bio}, network science \cite{network}, and social computing \cite{computing}. Formally, given an undirected graph $G = (V, E)$, a \emph{clique} is a subset of vertices $C \subseteq V$ such that every pair of vertices in $C$ is connected by an edge. The goal of the MCP is to find the maximum clique in a given graph $G$, which is defined as a fully connected subgraph with the largest number of vertices.

Due to its inherent combinatorial complexity, a wide variety of exact algorithms have been developed for MCP \cite{mcp}, focusing on different approaches to efficiently explore the solution space. Among these, the branch-and-bound and backtracking search are widely acknowledged for their ability to explore all possible cliques while pruning infeasible paths \cite{bab}. These methods are based on recursive strategies that break the problem down into smaller subproblems, progressively refining the search for the largest clique. Another important class of algorithms uses dynamic programming techniques that take advantage of overlapping subproblems to store intermediate results and avoid redundant calculations. Notably, these algorithms demonstrate \textbf{complementary performance profiles}—each tends to perform well on certain types of graphs but poorly on others. For example:

\begin{itemize}
    \item Algorithm LMC \cite{LMC} is suitable for handling graph instances with large maximum cliques, high graph density, or complex connectivity patterns;
    \item Algorithm dOmega \cite{dOmega} performs well on sparse graphs and those with community structures, but its runtime increases significantly for graphs with large clique-core gaps;
    \item Algorithm CliSAT \cite{CliSAT} specializes in highly dense, challenging graphs (e.g., those with frequent hidden structures and interference edges), though it shows weaker performance on regular graphs.
\end{itemize}

\begin{figure}[t]
\centering
\includegraphics[width=\linewidth]{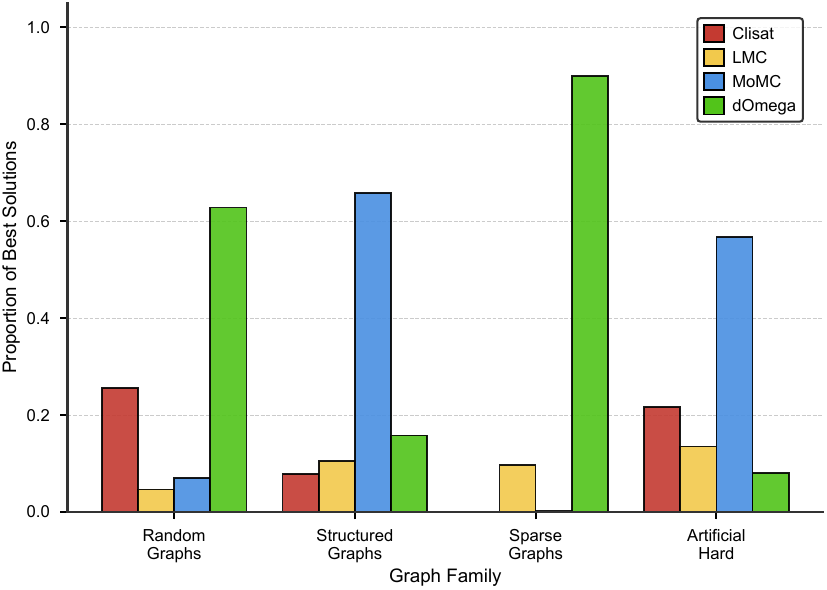}
\caption{Win-rate of four state-of-the-art solvers across graph families.}
\label{fig:motivation}
\end{figure}

Despite extensive algorithmic development for MCP, our observations indicate that no single solver is universally optimal. 
To illustrate this, we conducted a preliminary comparative study on diverse graph instances from the DIMACS benchmark and real-world networks (experimental settings are provided in Section \ref{sec:experiments}). 
The instances are grouped into four distinct families: Random Graphs (e.g., DSJC, p-hat, G-series), Structured Combinatorial Graphs (e.g., hamming, johnson, keller, MANN), Sparse Graphs (e.g., p2p, Maragal, tols), and Artificial Hard (e.g., brock, san, c-fat, gen). 
Four state-of-the-art exact MCP algorithms are evaluated: CliSAT \cite{CliSAT}, LMC \cite{LMC}, MoMC \cite{MoMC}, and dOmega \cite{dOmega}. 
The results, summarized in Fig.~\ref{fig:motivation}, show the proportion of instances within each graph family where an algorithm outperforms its competitors (comparisons are based first on the clique size found, with runtime used as a tie-breaker when sizes coincide).

Although CliSAT emerges as the best-performing solver overall in terms of average performance under the time limit (see Section \ref{sec:experimental_results}), Fig.~\ref{fig:motivation} reveals a far more nuanced picture.
As shown, the performance gaps reflect the intrinsic design biases of the solvers.
CliSAT, which integrates partial MaxSAT reasoning with colouring-based bounds and constraint propagation, captures 25.6\% of \emph{Random\_graphs} where these components effectively prune the search space on dense instances; yet the same mechanisms offer little advantage on \emph{Sparse\_graphs}, where it fails to win any instance.
dOmega dominates both \emph{Sparse\_graphs} (90.0\%) and \emph{Random\_graphs} (62.8\%).
On sparse instances its strength stems from low degeneracy and narrow clique-core gaps; on random graphs, which here include lower-density G-series instances, its degeneracy-based parameterization and exploitation of $k$-core structure allow it to handle moderate sizes and regular degree distributions well.
However, its advantage fades markedly where the clique-core gap widens: on \emph{Structured\_graphs} (15.8\%) and \emph{Artificial\_hard\_graphs} (8.1\%) it frequently falls behind and suffers timeouts.
MoMC, which employs a hybrid branching strategy that combines dynamic and static vertex selection with incremental MaxSAT reasoning, proves the most consistent solver on \emph{Structured\_graphs} (65.8\%) and \emph{Artificial\_hard\_graphs} (56.8\%), where reducing branching vertices is particularly beneficial.
LMC, built on core-number reduction, preprocessing, and incremental MaxSAT reasoning, attains moderate shares on \emph{Artificial\_hard\_graphs} (13.5\%) and \emph{Structured\_graphs} (10.5\%), yet it leads no family and remains below 5\% on \emph{Random\_graphs}.
While dOmega wins the largest number of instances overall owing to the heavy representation of sparse real-world graphs in the benchmark, its sharp decline on structured and hard instances confirms that no single solver can be trusted across all families.
Collectively, these family-specific patterns highlight the urgent need for instance-aware algorithm selection.

Building on these observations, we formalize the algorithm selection problem for MCP as follows. 
Although this paradigm has gained traction in domains such as SAT solving and the traveling salesman problem \cite{tsp_select}, MCP remains underserved by such efforts despite its practical relevance and algorithmic diversity.

To address this gap, we propose a novel two-stage methodology tailored to the algorithm selection problem for the MCP:

\begin{enumerate}
    \item \textbf{Instance-level feature representation}: We construct a labeled dataset by collecting diverse graph instances and extracting both global statistical features (e.g., number of nodes, edge count, density, assortativity) and local structural descriptors (e.g., node degrees, $k$-core values). These features serve as compact representations of the topological complexity of each instance.
    
    \item \textbf{Learning-based algorithm selection}: We train predictive models to associate the representations of instances with the most suitable exact algorithm. This process includes benchmarking four classical machine learning (ML) algorithms, as well as introducing a graph neural network (GNN)-based model.
\end{enumerate}

Traditional ML approaches face inherent limitations in this context. Most classical ML models rely primarily on manually crafted global graph features, which struggle to capture the intricate local topological structures (e.g., k-core neighborhoods, edge connectivity patterns). These structural factors are critical for determining which algorithm will perform best on a given graph instance. For instance, algorithms optimized for densely connected subgraphs (like those with prominent k-core structures) may outperform others on such instances \cite{xu2024efficient}, while edge connectivity patterns often dictate the efficiency of algorithms designed for sparse or modular graphs \cite{ye2023efficient}. Consequently, the predictive accuracy and generalization ability of traditional ML models degrade when applied to heterogeneous graph instances, such as dense community-structured graphs, or networks with highly skewed degree distributions (e.g., scale-free networks), where global statistical features struggle to adapt well to diverse topologies. GNNs, by contrast, address these limitations by systematically leveraging the rich relational information embedded in graph topology. Through neighborhood aggregation and message passing, GNNs can automatically learn discriminative local representations that complement global statistical features (e.g., average degree, clustering coefficient), thereby offering a more robust foundation for instance-aware algorithm selection.

Building on these insights, our core contribution lies in the latter stage: we design a dual-channel fusion architecture named GAT-MLP. This model explicitly decouples global and local information by integrating:
\begin{itemize}
    \item A \textbf{Graph Attention Network (GAT)} to encode local neighborhood patterns from raw graph topology
    \item A \textbf{Multilayer Perceptron (MLP)} to process global statistical features
\end{itemize}

By combining attention-driven topological encoding with high-level statistical signals, the dual-channel model achieves superior generalization across heterogeneous graph instances. Notably, our framework also brings dramatic gains in practical efficiency: as verified in Section~\ref{sec:experimental_results}, it introduces negligible inference overhead while achieving an up to $9.4\times$ speedup over the Single Best Solver (SBS) in end-to-end runtime.

The remainder of this paper is organized as follows. Section~\ref{sec:related_work} reviews the relevant work on exact MCP algorithms and algorithm selection techniques. Section~\ref{sec:method} presents our methodology, including feature extraction and the proposed dual-channel prediction framework that integrates GAT and MLP modules. Section~\ref{sec:experiments} details our experimental setup, evaluation results, and ablation studies. Finally, Section~\ref{sec:conclusion} concludes the paper and outlines future research directions.

\section{Related Work}
\label{sec:related_work}

\subsection{Exact Algorithms for the Maximum Clique Problem}

The Maximum Clique Problem, as a classic NP-hard problem, has been extensively studied, and numerous exact algorithms have been proposed. One of the earliest and most influential methods is the branch-and-bound algorithm introduced by Babel and Tinhofer in 1990 \cite{babel1990branch}, which utilizes graph coloring to determine upper bounds for the clique number. This algorithm laid the foundation for many subsequent developments in the field.

Subsequent work focused on improving pruning strategies to reduce the exponential search space. A well-known example is MaxCLQ \cite{maxclq}, which encodes MCP as a MaxSAT problem and utilizes MaxSAT reasoning to compute more stringent upper bounds. This method is particularly effective for handling dense graphs or graphs with irregular and complex structures. Based on this, BB-MaxClique \cite{BBMC} enhances the branch-and-bound framework with global vertex ordering and tighter upper bounds based on approximate coloring. Its efficient pruning and compact design offer strong performance in medium- to high-density graphs, making it a reliable baseline to evaluate modern clique algorithms.In 2015, Rossi et al. proposed the RGG algorithm \cite{RGG}, a parallel exact method that combines a fast core-number–based heuristic with tight upper bounds and aggressive pruning strategies. Designed for large-scale and sparse real-world networks, RGG achieves near-linear scalability and has been applied successfully to massive social and information graphs, making it a strong reference point for evaluating modern clique algorithms.

Although the above methods perform well in certain scenarios, they may still face efficiency bottlenecks in some graph instances. In recent years, researchers have proposed a series of more efficient maximum clique algorithms, including:

\begin{itemize}
    \item \textbf{dOmega:} This algorithm introduces a fixed parameter tractable (FPT) approach, with polynomial time complexity in terms of the graph size, but exponential complexity concerning the clique-core gap \cite{dOmega}. The clique-core gap is defined as:
\begin{equation}
    g = (d + 1) - \omega, 
    \label{eq:degeneracy_clique}
\end{equation}
    where \( \omega \) denotes the clique number of the graph and \( d \) is the \textit{degeneracy}, also known as the $k$-core number. Degeneracy is defined as the smallest integer \( d \) such that every subgraph of the graph contains at least one vertex of degree at most \( d \). It effectively measures the sparsity of the graph and serves as an upper bound on the size of its core substructures.The algorithm constructs candidate cliques starting from low-degree nodes and rapidly approaches the optimal solution by exploiting the relationship between degeneracy and the maximum clique size. Experimental results show that dOmega achieves high computational efficiency in sparse graphs and real-world social networks, particularly in scenarios where the degeneracy is low.

    \item \textbf{LMC:} This algorithm combines multilevel graph coloring and backtracking strategies to solve MCP \cite{LMC}. Within the traditional backtracking framework, LMC introduces multilevel coloring to dynamically compute the upper bound of each layer's candidate set and employs a greedy graph coloring algorithm to optimize pruning during the search. LMC uses local graph coloring information in each recursive layer, enabling fast pruning and efficient searching. This enhances its adaptability to graph structures, making it suitable for graphs with large maximum cliques, high graph density, or complex connectivity patterns.

    \item \textbf{MoMC:} MoMC is a maximum clique algorithm that dynamically adjusts the expansion order during the search process \cite{MoMC}. Select the best expansion order in real time based on node degree, color class size, and heuristic evaluations, thereby reducing the size of the search tree. The core innovation of MoMC lies in its dynamic sorting, which allows adaptive adjustment of the node expansion order based on the current search state, improving search efficiency. Furthermore, MoMC combines graph coloring and degree estimation to provide more accurate upper bounds for pruning, thus further enhancing search efficiency. MoMC is capable of handling large-scale dense graphs or irregularly structured graphs in a shorter time.

    \item \textbf{CliSAT:} CliSAT is an exact algorithm that integrates SAT solving techniques with MCP \cite{CliSAT}. By encoding MCP as a MaxSAT problem and using the MaxSAT reasoning mechanism, it improves the upper bound estimation and effectively prunes the search tree. Building upon traditional branch-and-bound algorithms, CliSAT incorporates SAT propagation and conflict detection mechanisms, allowing for more accurate solution space estimation. This is especially useful in high-density and difficult-to-solve graph instances. The key contribution of CliSAT is its ability to accelerate the pruning process using a MaxSAT solver, significantly improving solving efficiency without having to fully traverse the solution space. It has achieved outstanding results, particularly in large datasets and standard benchmark tests such as DIMACS.
\end{itemize}

These four algorithms leverage graph structures and modern solver technologies, outperforming traditional methods in multiple international benchmark tests, while exhibiting complementary performance profiles. Therefore, we have selected them for our experiments.

\subsection{Algorithm Selection for Combinatorial Problems}
The algorithm selection (AS) problem, originally formalized by Rice \cite{rice1976algorithm}, addresses how to choose the algorithm that performs best for a given problem instance based on its characteristics. In the context of combinatorial problems such as the Traveling Salesman Problem (TSP), the early approaches relied on hand-crafted features combined with machine learning classifiers \cite{tsp_select}. Seiler et al. \cite{seiler2020deep} proposed converting TSP instances into multiple image representations and using convolutional neural networks (CNN) for algorithm selection. Zhao et al. \cite{zhao2021towards} further advanced this approach with CTAS, a density-based CNN model that captures the density features of problem instances. More recently, Garouani et al. \cite{garouani2025experimental} reviewed meta-learning approaches for automated algorithm selection, providing valuable insights into current research trends. Kostovska et al. \cite{kostovska2023comparing} compared multiple AS methods for black-box optimization problems, showing the strengths and weaknesses of statistical and deep learning models. Furthermore, Liao et al. \cite{liao2025bopo} proposed a neural preference optimization method that replaces reinforcement learning with pairwise ranking objectives, which has found promising applications in TSP and job-shop scheduling. Wu et al. \cite{wu2023large} proposed an LLM-enhanced AS strategy encoding problem instances and algorithm metadata for zero/few-shot selection, but it does not target graph-structured problems. Song et al. \cite{song2023revisit} introduced GINES, a GIN-based GNN for node aggregation, which relies solely on injective neighbourhood aggregation.

Among mainstream graph neural networks for graph-based combinatorial optimization tasks, GCN \cite{gcn} and GIN \cite{gin} are classic models built on static neighborhood aggregation mechanisms, which have been widely adopted for learning graph-structured problem instance features. GraphSAGE (SAGE) \cite{sage}, as an inductive learning framework, leverages neighbor sampling to enable scalable graph representation learning, making it suitable for large-scale combinatorial optimization scenarios. These representative models and their enhanced variants will be included as key baselines in our experimental evaluation, where we conduct comprehensive comparative analyses to validate the performance of our proposed approach (see Section \ref{sec:baselines}).

\section{Methodology}
\label{sec:method}
\begin{figure*}[t]
    \centering
    \includegraphics[width=0.9\linewidth]{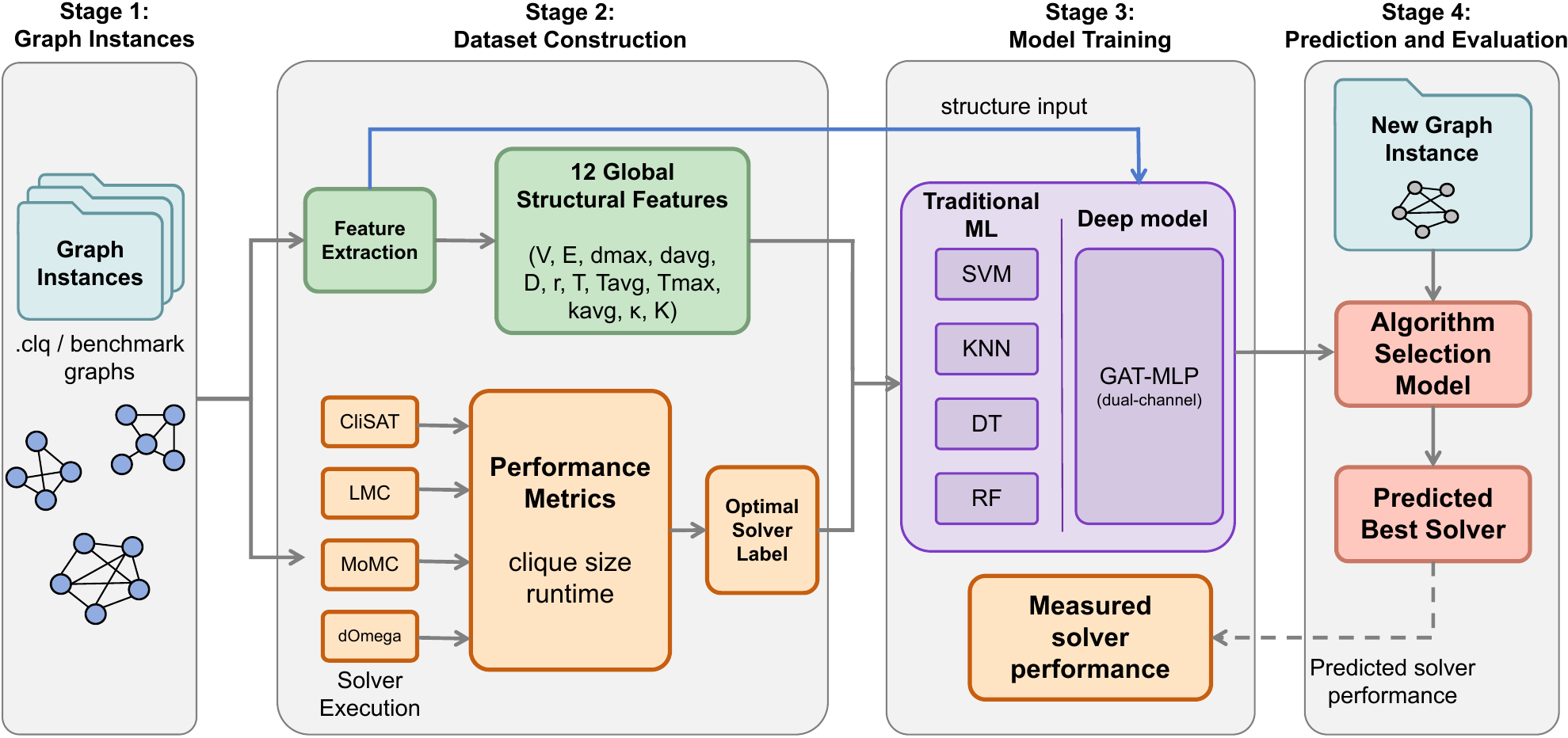}
    \caption{The proposed framework for algorithm selection in the MCP.}
    \label{fig:work}
\end{figure*}

We propose a framework for algorithm selection in the MCP using machine learning techniques, as illustrated in Fig.~\ref{fig:work}. 

In the dataset preparation step, a collection of graph instances is provided. From these graph instances, structural features are extracted, and four candidate MCP algorithms are executed to obtain their performance metrics. Based on these metrics, the optimal algorithm can be identified for each instance. Furthermore, the collected performance data serves as an evaluation criterion for the trained model.

During the model training step, the extracted features and algorithm performance metrics are jointly used to train several machine learning models, including Support Vector Machine (SVM), K-Nearest Neighbors (KNN), Decision Tree (DT), Random Forest (RF), and a dual-channel GAT-MLP. The GAT-MLP allows the model to leverage graph structural information to improve the accuracy of the prediction. When a new graph instance is provided, its features are extracted and fed into the trained model, which predicts the most suitable algorithm for solving the given MCP instance. The following subsections provide detailed descriptions of these steps and the dual-channel architecture of the proposed GAT-MLP.

\subsection{Dataset Preparation and Classical Machine Learning Models}
\label{sec:dataset_preparation}
For each graph instance, 12 structural features are computed to capture key topological properties of the graph, serving as inputs for the algorithm selection model. Table \ref{tab:features} presents the 12 features. In previous studies on the maximum clique problem, factors such as graph density, degree distribution, and core structure, among others, have been repeatedly shown to exert a decisive influence on algorithmic performance \cite{mcp}. Therefore, selecting these 12 structural features allows us to capture the key topological properties known to affect solver behavior. Among these, the assortativity coefficient quantifies the tendency of nodes to connect with others that have similar degrees, reflecting the structural organization of the overall network \cite{newman2002assortative}. For undirected graphs, it is calculated as the Pearson correlation coefficient between the degrees of nodes at either end of each edge, as follows:
\begin{equation} r = \frac{ \frac{1}{M} \sum_i j_i k_i - \left( \frac{1}{2M} \sum_i (j_i + k_i) \right)^2 }{ \frac{1}{2M} \sum_i (j_i^2 + k_i^2) - \left( \frac{1}{2M} \sum_i (j_i + k_i) \right)^2 } \label{eq:ac}, 
\end{equation}
where \( M \) is the total number of edges in the graph, and \( j_i \) and \( k_i \) denote the degrees of the two nodes connected by the \( i \)-th edge. The clustering coefficient (denoted as \( C \)) quantifies how likely neighbors of a node are to be interconnected, indicating the density of local clusters. The local clustering coefficient is calculated for each node \(i\) as:
\begin{equation}
C_i = \frac{2 e_i}{k_i(k_i - 1)}, 
\label{eq:cc}
\end{equation}
where \( e_i \) is the number of edges between the neighbors of node \( i \), and \( k_i \) is the degree of node \( i \). The global clustering coefficient (\( \kappa \)) characterizes the overall clustering tendency of the entire network:
\begin{equation}
\kappa = \frac{3 T}{N_T}, 
\label{eq:gcc}
\end{equation}
where \( T_G \) is the number of triangles in the network, and \( N_T \) is the number of connected triplets of nodes (triplets of nodes where at least two are adjacent). The $k$-core identifies the largest subgraph where every node has at least $k$ neighbors, highlighting densely connected regions within the graph.

Covering global topology, local node interactions, and subgraph-level properties, these 12 structural features provide a comprehensive and discriminative characterization of MCP instances.

\begin{table}[t]
\renewcommand{\arraystretch}{1.6}
\centering
\caption{Twelve graph structural features and their descriptions.}
\label{tab:features}
\begin{tabularx}{\columnwidth}{l X}
\toprule
\textbf{Symbol} & \textbf{Description} \\
\midrule
$V$ & Total number of nodes \\
$E$ & Total number of edges \\
$d_{\max}$ & Maximum degree \\
$d_{\mathrm{avg}}$ & Average degree \\
$D$ & Graph density \\
$r$ & Assortativity coefficient \\
$T$ & Total number of triangles (3-cliques) in the graph \\
$T_{\mathrm{avg}}$ & Average number of triangles formed by an edge \\
$T_{\max}$ & Maximum number of triangles formed by an edge \\
$\kappa_{\mathrm{avg}}$ & Average local clustering coefficient \\
$\kappa$ & Global clustering coefficient \\
$K$ & Maximum $k$-core number \\
\bottomrule
\end{tabularx}
\end{table}

We apply four exact algorithms to solve the MCP on each graph instance: \textbf{Clisat}, \textbf{LMC}, \textbf{MoMC}, and \textbf{dOmega}. Each algorithm returns the size of the maximum clique it finds along with the corresponding computational time. To determine the optimal algorithm for each instance, we adopt the following criteria: the algorithm that identifies the largest clique is selected; if multiple algorithms return the same clique size, the one with the shortest computational time is preferred.

\begin{figure}[t]
    \centering
    \includegraphics[width=0.95\linewidth]{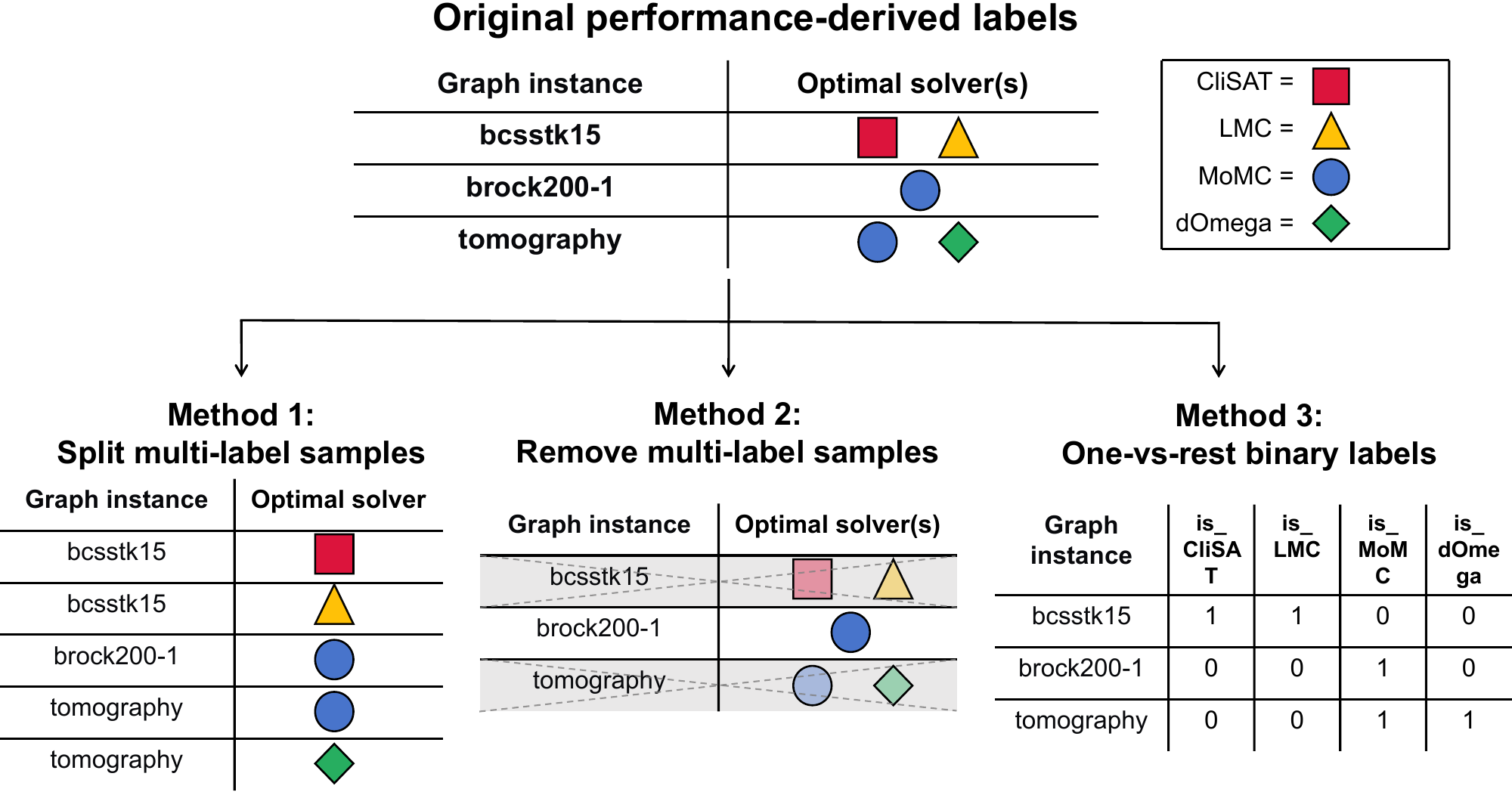}
    \caption{Overview of the three methods for handling multi-label samples.}
    \label{fig:multi_label_methods}
\end{figure}

In rare cases, multiple algorithms produce identical results on the same graph instance, resulting in multi-label instances. Inspired by the work of Kanda et al.~\cite{kanda2011selection}, we investigated three methods to handle such scenarios, as illustrated in Fig.~\ref{fig:multi_label_methods}.

\begin{enumerate}
    \item \textbf{Method 1: Splitting multi-label instances into multiple single-label instances.} Each multi-label instance is divided into several single-label instances, where the number of resulting instances equals the number of associated labels. A notable drawback of this approach is that it generates instances with identical features but different labels.
    
    \item \textbf{Method 2: Removing multi-label instances.} All graph instances associated with more than one optimal algorithm are excluded from the dataset, simplifying the classification task. However, this method may lead to the loss of potentially valuable training data.
    
    \item \textbf{Method 3: Training independent binary classifiers for each label.} A separate binary classifier is trained for each label, treating the target label as the positive class and all the others as the negative class. The predictions from these classifiers are then combined to produce the final multi-label prediction. However, this method may suffer from class imbalance as the positive class is typically much smaller than the negative class.
\end{enumerate}

These three methods produce three distinct datasets. These datasets are then used to train and test four machine learning classifiers: RF, DT, KNN and SVM, respectively.

After training and testing the models, we compare the performance of each classifier on the three processed datasets.  The evaluation considers key performance indicators such as accuracy and F1 scores. Then we select the optimal combination of method and classifier based on their performance across these evaluation metrics.  The goal is to identify a model that not only achieves accurate clique prediction but also ensures computational efficiency.

\subsection{Dual-Channel Architecture}

\begin{figure*}[t]
  \centering
  \includegraphics[width=0.95\linewidth]{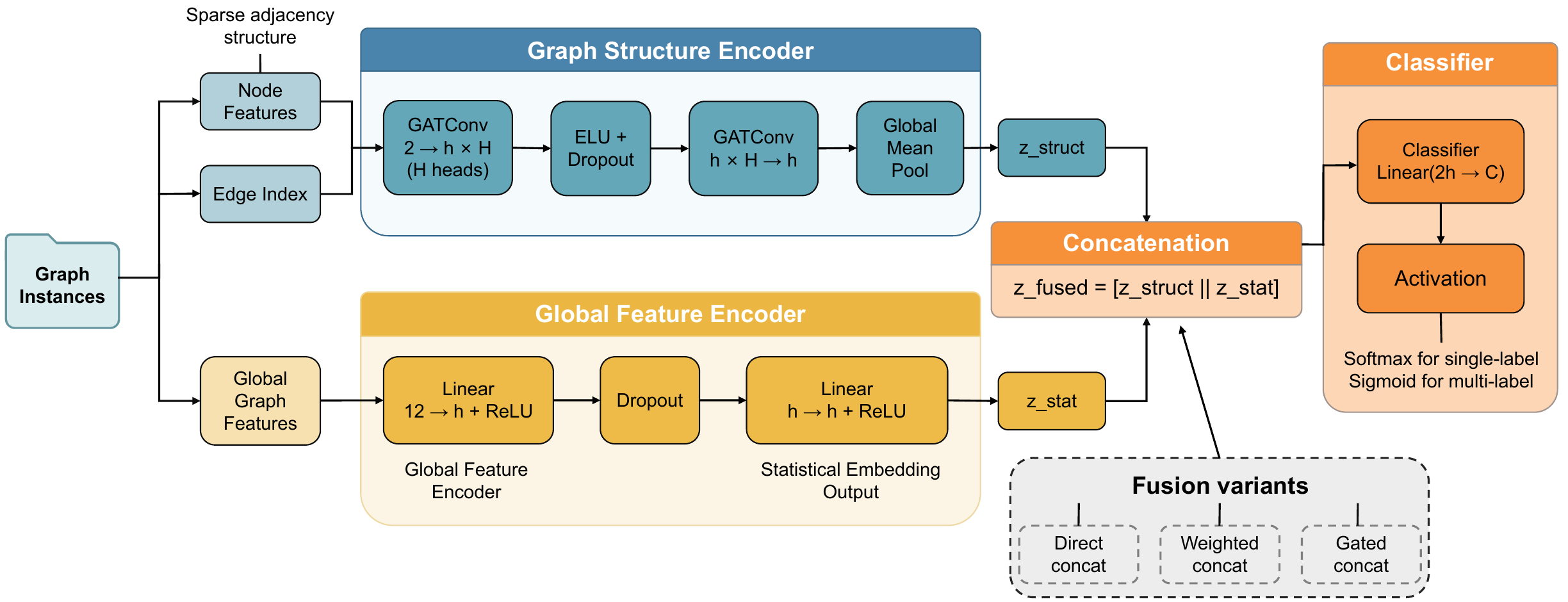}
  \caption{Architecture of the proposed dual-channel GAT model, consisting of a structure encoder (graph-based) and a statistical encoder (global features), followed by feature fusion and classification.}
  \label{fig:dual_gat}
\end{figure*}
We propose a dual-channel architecture (see Fig.~\ref{fig:dual_gat}) that integrates a structure encoder for topological feature extraction and a statistical encoder for global distribution modeling, enabling comprehensive graph representation learning through feature fusion.

\textbf{Graph Structure Encoder.}
This branch operates directly on the graph topology encoded in the \texttt{.clq} files. Each graph $G = (V, E)$ is represented as an undirected graph with node set $V$ and edge set $E$. For each node $v \in V$, we assign a two-dimensional initial feature $x_v \in \mathbb{R}^2$ representing its degree and core number. These features are normalized using min-max scaling. The normalized features are passed through a two-layer GAT~\cite{velickovic2017graph}, which performs neighborhood-based message passing using learned attention weights. The first GATConv layer transforms the node features from $\mathbb{R}^{n \times 2}$ to $\mathbb{R}^{n \times (h \cdot H)}$, where $h$ is the hidden size and $H$ is the number of attention heads. The second GATConv layer aggregates the multi-head outputs and maps them to $\mathbb{R}^{n \times h}$. To obtain a graph-level representation, we apply a global mean pooling operation across all node embeddings, resulting in a fixed-length vector $\mathbf{z}_\text{struct} \in \mathbb{R}^h$.

\textbf{Global Feature Encoder.}
This branch encodes global graph properties using 12 global statistical features extracted from each instance. These features capture diverse structural statistics, including node and edge counts, degree-based measures, graph density, clustering coefficients, triangle-related metrics, and core numbers. The complete list and descriptions of these features are presented in Table~\ref{tab:features}. These features are assembled into a vector $\mathbf{g} \in \mathbb{R}^{12}$, normalized using z-score normalization. The normalized vector is then fed into a MLP consisting of two linear layers with ReLU activations. Dropout is applied between layers to prevent overfitting. This encoder transforms the input as $\mathbb{R}^{12} \rightarrow \mathbb{R}^h \rightarrow \mathbb{R}^h$, producing a global embedding vector $\mathbf{z}_\text{stat} \in \mathbb{R}^h$.

\textbf{Fusion and Classifier.}
We design three candidate connection strategies for this fusion step, and adopt direct concatenation as the primary approach in this work—one where the two embeddings are combined without additional parameterization. Specifically, the two embeddings $\mathbf{z}_\text{struct}$ and $\mathbf{z}_\text{stat}$ are concatenated to form a joint representation $\mathbf{z}_\text{fused} = [\mathbf{z}_\text{struct} \| \mathbf{z}_\text{stat}] \in \mathbb{R}^{2h}$, which captures local and global structural information.

The other two alternative strategies are reserved for detailed analysis in Section~\ref{sec:ablation_study} to quantify their impact on model performance:
\begin{enumerate}
    \item \textbf{Weighted Concatenation}: Introduces learnable parameter vectors to adaptively assign weights to $\mathbf{z}_\text{struct}$ and $\mathbf{z}_\text{stat}$ before concatenation;
    \item \textbf{Gated Concatenation}: Uses a gating network with sigmoid activation to dynamically adjust the contribution ratio of the two embeddings during fusion.
\end{enumerate}

The fused vector is passed to a final linear layer that projects it to $\mathbb{R}^C$, where $C$ is the number of output classes. During inference, we apply a softmax (for single-label classification) or a sigmoid (for multilabel classification) activation externally to obtain probabilities. An early stopping strategy based on validation performance is employed to prevent overfitting.

This dual-channel design is elaborately tailored for MCP algorithm selection, rather than a naive combination of general modules. The graph structure encoder focuses on capturing core-related local topology, while the global feature encoder specializes in modeling holistic graph statistics, exhibiting clear functional alignment with the structural properties critical to MCP solvers. The concatenation‐based fusion strategy ensures effective interaction between multi-granularity representations.

The necessity of the two-branch framework and the effectiveness of different fusion strategies are systematically verified via ablation studies in Section~\ref{sec:ablation_study}. This purpose‐built design enables the model to leverage fine-grained local interactions within the graph while incorporating global structural characteristics, yielding more robust and generalizable performance for MCP algorithm selection.

\subsection{Baselines}
\label{sec:baselines}

To comprehensively assess the contribution of global information, we select three widely recognized graph neural networks, GCN~\cite{gcn}, GIN~\cite{gin,song2023revisit}, and SAGE~\cite{sage}, as our initial baselines. These models cover representative message passing paradigms and serve as a solid foundation for isolating the impact of global cues. However, since standard architectures primarily rely on node-level attributes and local neighbourhood aggregation, their ability to exploit global context is inherently constrained. To address this gap, we further extend each baseline with two forms of global-aware enhancement. The late-fusion variants (denoted as "Global") let the graph encoder and a lightweight MLP digest the structure and the global statistics separately. And a learnable convex gate then decides how much weight each stream deserves. In contrast, the early-fusion variants (denoted as "Global Feature Enhancement Network (GFEN)") compress the global vector to a scalar and concatenate it directly with the pooled graph embedding before the final classifier, so the decision layer sees both cues at once. All baselines and their enhanced counterparts are trained under the same data splits, early-stopping protocol and random seeds, yielding a fair spectrum of comparisons ranging from purely structural to fully global-contextualized models.

\section{Experiments}
\label{sec:experiments}

\subsection{Experiment Settings}
\label{sec:exp-settings}

\subsubsection{Experimental Environment}
We conducted our experiments in two different computing environments for practical and efficiency considerations. The first environment is a high-performance server, used to run exact MCP algorithms on graph instances. Since exact algorithms require an extensive computational time when dealing with challenging instances, a more powerful computational resource is necessary. This server is equipped with an Intel(R) Xeon(R) Silver 4210R CPU @ 2.40GHz (20 cores), running Ubuntu Linux 22.04 with kernel version \texttt{6.8.0-60-generic}. The second environment is used to train and evaluate graph learning models. It has pre-installed PyTorch Geometric, NetworkX, and other required Python dependencies. This setup is well suited for deep learning development without additional configuration. It features a 13th Gen Intel(R) Core(TM) i7-13650HX processor with a base frequency of 2.60 GHz.

\subsubsection{Datasets}
\label{sec:datasets}
Due to the lack of standardized datasets and limited prior work on data-driven approaches for algorithm selection in MCP, we constructed a custom dataset to support our learning-based study. Graph instances are collected from two publicly available sources: the Network Repository\footnote{\url{https://networkrepository.com/}} and the CSPLIB library~\cite{csplib}. From the Network Repository, we include graphs from the well-known DIMACS and BHOSLIB benchmarks, which are widely used in MCP research. Additionally, to enhance sample diversity and quantity, we incorporate selected undirected and unweighted instances from the "Miscellaneous Networks" category.

Although the Network Repository provides certain structural features for each graph, many instances have incomplete data or values with limited numerical precision. To ensure consistency and accuracy, we recompute all twelve global features listed in Table~\ref{tab:features} directly from the raw graph data. These features are used as input to the global feature encoder branch of our model.

To obtain the target labels, we run four exact MCP algorithms on each graph. A time limit of 21,600 seconds (6 hours) is applied to most instances, while a shorter limit of 1,800 seconds (0.5 hours) is used for graphs from DIMACS and CSPLIB to ensure computational feasibility. After filtering out unsolved graphs (those exceeding the time limit) and trivial cases, we retain a total of 572 labeled graph instances.

The resulting dataset spans a wide range of graph sizes and structural complexity, covering small- to large-scale graphs. The number of nodes ranges from 28 to 55,371, and the number of edges ranges from 210 to 568,960. Graph densities vary from extremely sparse (less than 0.001) to highly dense (up to 0.9963), ensuring the inclusion of both sparse and dense graph scenarios. This structural diversity provides a challenging benchmark for MCP classification tasks.

\subsubsection{Evaluation Metrics}
We report classification accuracy along with both macro-averaged and weighted F1 scores (Macro-F1 and Weighted-F1) to evaluate model performance. While accuracy reflects the overall correctness of predictions, it may be biased in the presence of class imbalance. The F1 score, which considers both precision and recall, provides a more balanced assessment, especially in datasets with skewed class distributions \cite{sokolova2009systematic}.

Accuracy is defined as the ratio of correctly predicted instances to the total number of instances:
\begin{equation}
    \text{Accuracy} = \frac{TP + TN}{TP + TN + FP + FN}, 
\end{equation}
where $TP$, $TN$, $FP$, and $FN$ denote true positives, true negatives, false positives, and false negatives, respectively.

Since the task involves selecting among four candidate algorithms, it constitutes a multi-class classification problem. Therefore, we adopt Macro-F1 as a primary evaluation metric to account for performance across all classes. Macro-F1 calculates the F1 score for each class independently and then averages them, ensuring that each class contributes equally regardless of its frequency. The Macro-F1 score is defined as follows:
\begin{equation}
    \text{Macro-F1} = \frac{1}{C} \sum_{c=1}^{C} \text{F1}_c, 
\end{equation}
where \( C \) is the total number of classes, and the F1 score per class is defined as:
\begin{equation}
    \text{F1}_c = 2 \cdot \frac{\text{Precision}_c \cdot \text{Recall}_c}{\text{Precision}_c + \text{Recall}_c}, 
\end{equation}
with
\begin{equation}
    \text{Precision}_c = \frac{TP_c}{TP_c + FP_c}, \quad
    \text{Recall}_c = \frac{TP_c}{TP_c + FN_c}, 
\end{equation}
Here, \( TP_c \), \( FP_c \), and \( FN_c \) denote the true positives, false positives, and false negatives for class \( c \), respectively. Macro-F1 treats all classes equally, regardless of their frequency, making it especially suitable for imbalanced multi-class settings.

In addition, we also report the Weighted-F1, which takes into account class imbalance by weighting each class-specific F1 score according to the number of true instances in that class:
\begin{equation}
    \text{Weighted-F1} = \sum_{c=1}^{C} \frac{n_c}{N} \cdot \text{F1}_c, 
\end{equation}
where \( n_c \) is the number of true instances for class \( c \), and \( N = \sum_{c=1}^{C} n_c \) is the total number of instances. While Macro-F1 emphasizes equal importance for all classes, Weighted-F1 reflects the overall performance of the model by taking the class distribution into account.

\subsubsection{Hyperparameter Settings}

This work focuses on evaluating the overall performance of different methods on the MCP. All models are trained and tested on the same data split, with 80\% of the data used for training and 20\% for testing. For traditional ML models, we use standard numerical features as input and apply 5-fold cross-validation with \texttt{GridSearchCV} to select the optimal hyperparameters. For GAT-MLP, we use the Adam optimizer with a learning rate of 0.001 and a weight decay of $1 \times 10^{-4}$. The hidden dimension is set to 32, the dropout rate to 0.5, the batch size to 16, and training runs for up to 50 epochs. A detailed sensitivity analysis of the learning rate and dropout rate is provided in Section~\ref{sec:hyper_analysis}.

\subsection{Experimental Results}
\label{sec:experimental_results}

\begin{figure}[t]
    \centering
    \includegraphics[width=0.48\textwidth]{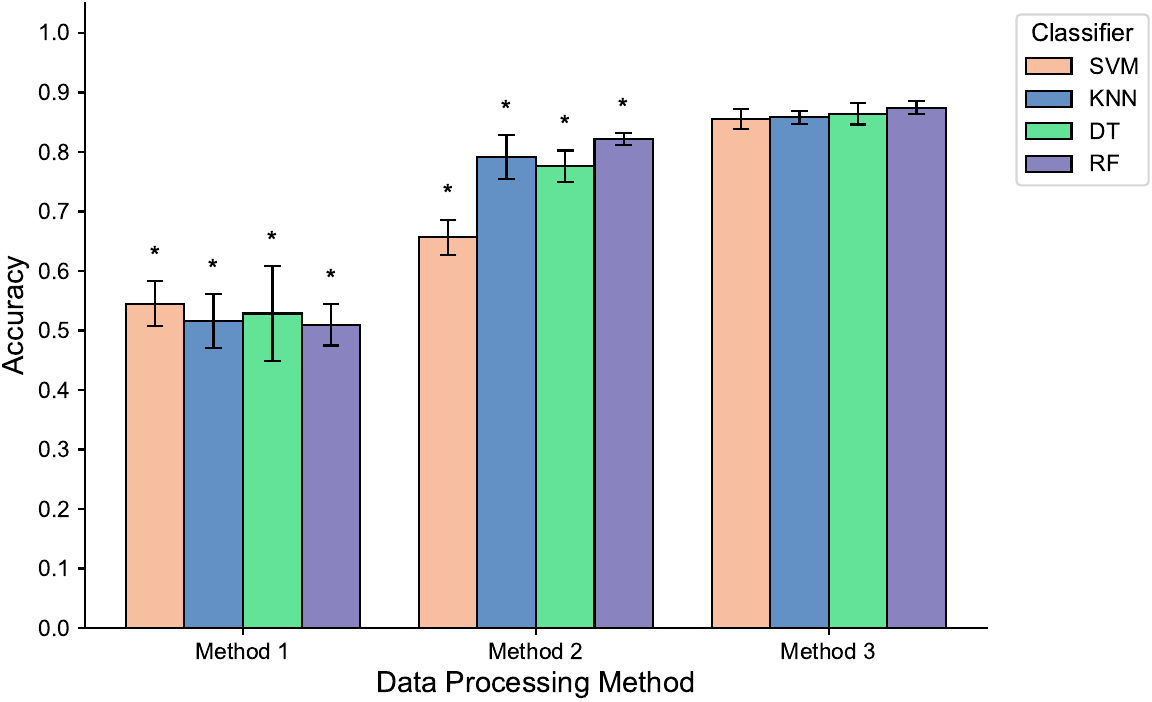}
    \caption{Accuracy of four classifiers under three feature-engineering methods (mean $\pm$ std; *: Mann--Whitney U test vs. best, $p<0.05$).}
    \label{fig:accuracy}
\end{figure}

We evaluated four classical machine learning models, including SVM, DT, RF and KNN, under three different feature engineering strategies (Method 1–3, see Section~\ref{sec:dataset_preparation}). Our primary objectives are to assess the relative merits of these partitioning strategies and to compare the performance of the four classifiers under each strategy. Classification accuracy across all configurations is shown in Fig.~\ref{fig:accuracy}. All reported metrics are presented as mean $\pm$ standard deviation (std).

\begin{table*}[t]
\renewcommand{\arraystretch}{1.25}
\centering
\caption{Feature importance matrix under Method 1 and Method 2 using DT and RF, where \textbf{Bold} indicates the most important feature, and \underline{underlined} denotes the second most important.}
\label{tab:importance_matrix}
\begin{tabularx}{\textwidth}{l *{4}{>{\centering\arraybackslash}X}}
\toprule
\textbf{Feature} 
& \multicolumn{2}{c}{\textbf{Method 1}} 
& \multicolumn{2}{c}{\textbf{Method 2}} \\
\cmidrule(lr){2-3} \cmidrule(lr){4-5}
& DT & RF & DT & RF \\
\midrule
$V$                & 0.0948 & 0.0971 & 0.0572 & 0.0777 \\
$E$                & 0.0274 & 0.0501 & 0.0557 & 0.0614 \\
$d_{\max}$         & 0.0161 & 0.0455 & 0.0555 & 0.0496 \\
$d_{\mathrm{avg}}$ & 0.0166 & 0.0919 & 0.0053 & 0.0767 \\
$D$                & \textbf{0.4487} & \textbf{0.1633} & \textbf{0.4970} & \textbf{0.1437} \\
$r$                & 0.0860 & 0.0699 & \underline{0.1331} & 0.0734 \\
$T$                & 0.0175 & 0.0529 & 0.0225 & 0.0522 \\
$T_{\mathrm{avg}}$ & 0.0528 & \underline{0.1090} & 0.0310 & 0.1175 \\
$T_{\max}$         & \underline{0.1120} & 0.0708 & 0.0678 & 0.0684 \\
$\kappa_{\mathrm{avg}}$ & 0.0184 & 0.0669 & 0.0481 & 0.0617 \\
$\kappa$           & 0.0234 & 0.0667 & 0.0217 & 0.0845 \\
$K$                & 0.0863 & 0.1160 & 0.0051 & \underline{0.1331} \\
\bottomrule
\end{tabularx}
\end{table*}

As shown in Fig.~\ref{fig:accuracy}, Method 1 yields markedly inferior performance. Even the best model under Method 1 (SVM) achieves only $54.45 \pm 3.78\%$, with the remaining classifiers performing similarly. This poor performance arises from the fact that identical feature vectors are assigned conflicting labels, effectively introducing severe label noise. In contrast, both Method 2 and Method 3 substantially improve accuracy for all classifiers. Tree-based models (DT and RF) exhibit particularly strong performance: under Method 2, RF attains the highest accuracy ($82.17 \pm 0.97\%$), and under Method 3 it further increases to $87.39 \pm 1.08\%$, with DT and KNN trailing only slightly. Overall, RF is the best-performing classifier across all methods.

\begin{figure*}[t]
    \centering
    \includegraphics[width=0.95\textwidth,height=0.28\textheight,
  keepaspectratio]{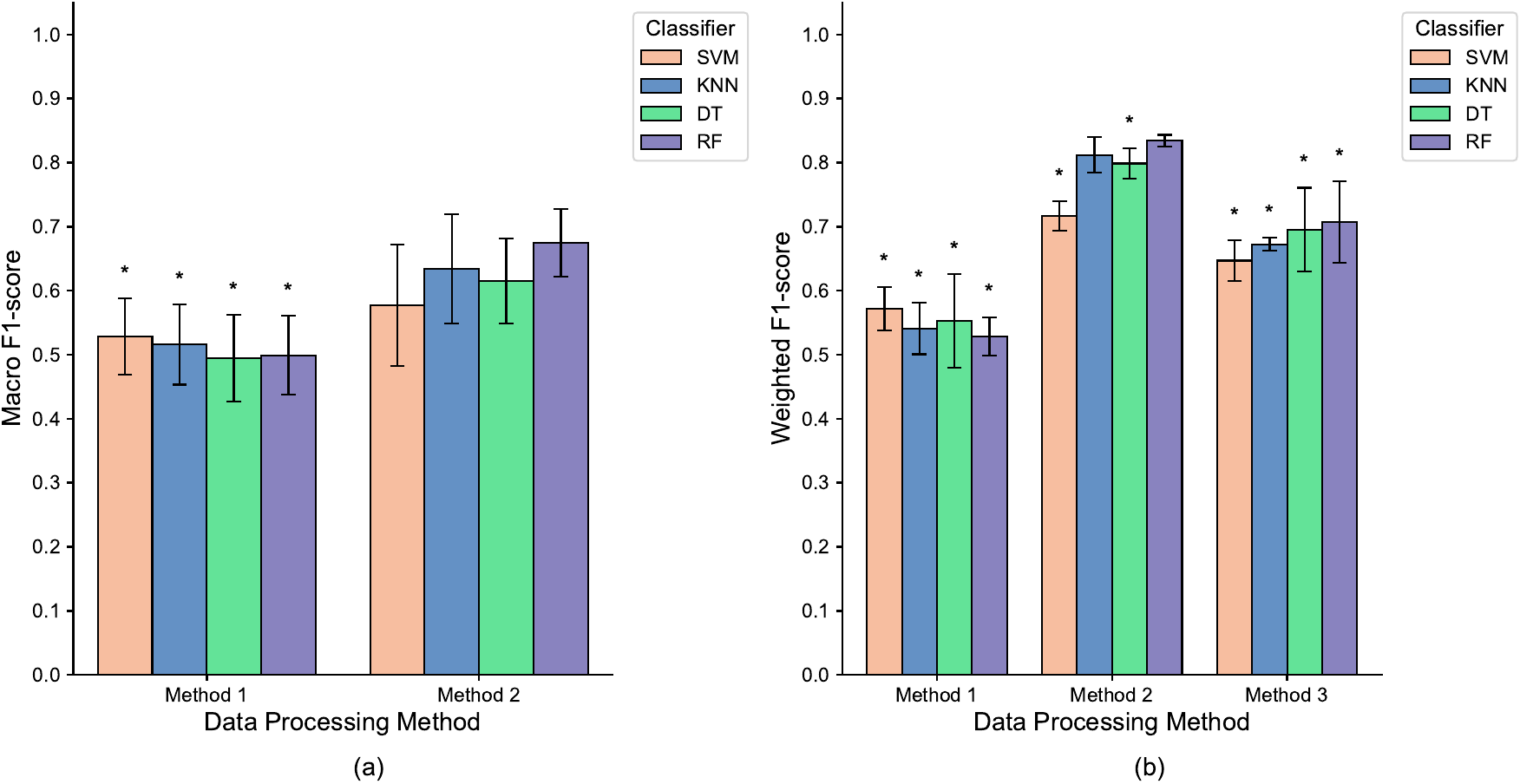}
    \caption{F1 comparison of four classifiers under different feature-engineering methods. (a) Macro-F1; (b) Weighted-F1 (mean $\pm$ std; *: Mann--Whitney U test vs. best, $p<0.05$).}
    \label{fig:bar}
\end{figure*}

F1-scores under the three methods are presented in Fig.~\ref{fig:bar}. For Method~1, both the Macro-F1 and Weighted-F1 remain below $0.56$ for all classifiers on average, consistent with its poor accuracy. Under Method~2, the F1-scores increase markedly: RF obtains the highest Weighted-F1 ($0.834 \pm 0.009$), with KNN and DT performing closely behind, while SVM yields the lowest Macro-F1, suggesting that its accuracy advantage originates primarily from majority-class bias. In Method 3, the accuracy and per-label F1-scores of tree-based models further improve compared with Method 2; however, the aggregated Weighted-F1 values are consistently lower than those of Method 2. This is because Method 3 converts the original multi-class problem into four independent one-vs-rest binary classification tasks (see Section~\ref{sec:method}), where each classifier focuses solely on its own target label. The resulting F1-scores therefore represent label-specific performance and already incorporate the imbalance between positive and negative samples within each binary task. Nevertheless, the severe class imbalance inherently limits performance. For example, the rarest label exhibits a positive-to-negative ratio of approximately $1{:}23$, making it difficult for the classifier to accurately recognize the minority class and ultimately lowering the aggregated F1 performance.

Overall, Method 1 proves to be ineffective for this dataset. The main issue lies in the artificial instances it generates, where feature vectors are identical but labels differ, which confuses the classifier’s decision boundary. Between the remaining two strategies, Method 3 achieves marginally higher accuracy, but at the cost of a lower Weighted-F1 and higher computational complexity due to the training of multiple binary classifiers under severe class imbalance. Method 2, on the other hand, achieves a more favorable balance between accuracy and F1-based metrics. Among the four classifiers, tree-based models, especially RF, consistently show strong performance across all evaluation metrics, highlighting their robustness under different data-partitioning strategies.

To explore how tree-based models make decisions, we analyzed feature importance for DT and RF under Method 1 and Method 2, using Gini impurity reduction to quantify importance. Table~\ref{tab:importance_matrix} presents the relative importance scores of the twelve structural features of the graph in both models and methods.

\begin{figure*}[t]
    \centering
    \includegraphics[width=0.75\textwidth]{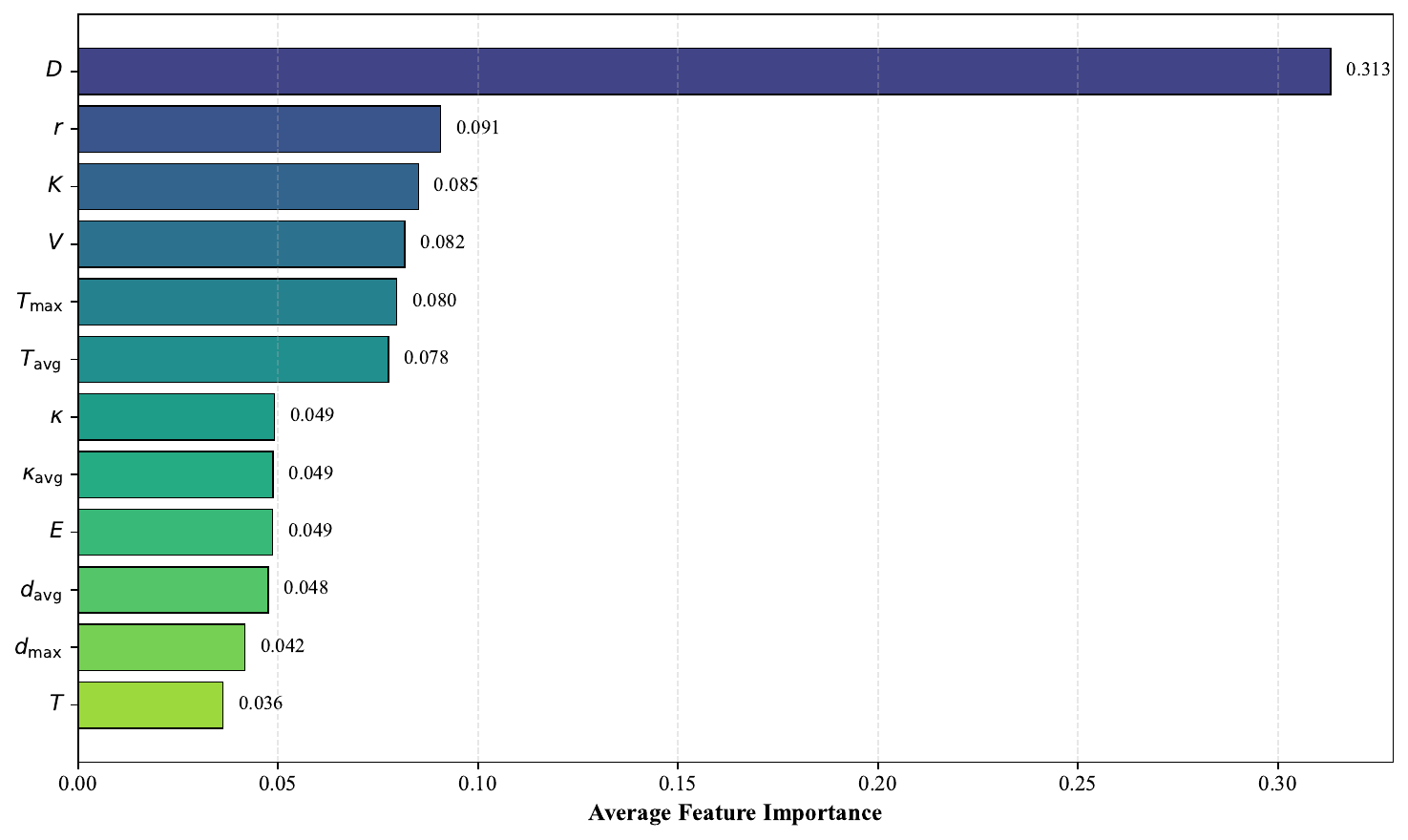}
    \caption{Average feature importance across Method 1/2 under DT and RF.}
    \label{fig:avg_imp}
\end{figure*}

Among all features, graph density $D$ consistently shows the highest importance across all models and experimental settings, highlighting the decisive role of global compactness and overall connectivity in differentiating MCP algorithm performance. We also compute the average importance across all configurations (DT and RF under Method 1 and Method 2), as visualized in Fig.~\ref{fig:avg_imp}. The dominant status of $D$ is clearly observed in all scenarios, reaffirming its core contribution to the classification task.

Apart from graph density, topological metrics reflecting holistic graph properties and core structural characteristics yield considerable predictive effects. The assortativity coefficient $r$, node count $V$, maximum $k$-core number $K$, and clique-related indicators (i.e., $T_{\mathrm{avg}}$, $T_{\max}$) occupy relatively high average importance. In contrast, local microscopic metrics such as average degree $d_{\mathrm{avg}}$ exhibit limited contribution individually. These distinctions indicate that macro-level global attributes and high-order core structures serve as more reliable discriminative cues for MCP solver selection, rather than simple local node-level topological statistics.

Interestingly, these feature importance patterns closely reflect the theoretical behaviors and structural preferences of different exact MCP algorithms discussed in Section~\ref{sec:related_work}. For example, \textbf{dOmega} is known to perform efficiently in sparse graphs with low degeneracy, which aligns with the importance of the maximum $k$-core number $K$, a key indicator of core-like structures and graph degeneracy. While the average degree $d_{\mathrm{avg}}$ individually contributes less, its inherent association with core structure when combined with $K$ still indirectly supports the structural adaptability of dOmega. In contrast, algorithms such as \textbf{LMC} and \textbf{MoMC}, which leverage graph coloring and adaptive expansion heuristics, benefit from dense connectivity and structural regularity. These conditions are captured by high values of $D$ (graph density) and clustering-related features like $\kappa$ and $T_{\mathrm{avg}}$, which are consistent with their reliance on structural regularity. Similarly, the effectiveness of \textbf{CliSAT} in dense and complex instances is reflected in its reliance on structural indicators such as density ($D$) and clique-related patterns ($T_{\mathrm{avg}}$, $T_{\max}$), reinforcing the ability of these features to distinguish between algorithms' performance.

These findings offer valuable insights that directly inform the design of our dual-channel GAT-MLP architecture. The observed dominance of global features such as graph density ($D$) and the relevance of core-related metrics (e.g., $K$) and holistic topological attributes (e.g., $r$, $V$) suggest that both global and local high-order structural properties play essential yet distinct roles in MCP classification. Motivated by this, we design a model that explicitly separates and specializes the processing of these two feature types: one branch focuses on global structural characteristics, and the other focuses on local core topology to fully capture the discriminative information required for MCP algorithm selection.

Lastly, we evaluate GAT-MLP, the dual-channel architecture presented in Section~\ref{sec:method} and illustrated in Fig.~\ref{fig:dual_gat}. To verify the contribution of this design, we compare GAT-MLP against both classical ML representatives (RF) and a spectrum of graph baselines (GCN, GIN, SAGE) together with their global-aware enhancements: late-fusion (\textit{-Global}) and early-fusion (\textit{-GFEN}) variants, detailed in Section~\ref{sec:baselines}.

While earlier comparisons of traditional ML models under different data splits employed the Mann--Whitney U test vs.\ the best performer, here we adopt the paired $t$-test because all neural models are evaluated under identical train/validation/test partitions and random seeds, yielding paired observations per run.

As summarized in Table~\ref{tab:baselines_compare}, GAT-MLP consistently achieves the highest performance: accuracy $\mathbf{90.43\%}\pm3.95$, Macro-F1 $\mathbf{0.690}\pm0.048$, and Weighted-F1 $\mathbf{0.892}\pm0.045$.
Notably, these scores exceed the strongest GNN-based competitor by $+3.04\%$ accuracy (GCN-GFEN), $+0.238$ Macro-F1 (GIN-Global), and $+0.065$ Weighted-F1 (GIN-Global). The substantial gains confirm that jointly modeling structural topology and statistical semantics is highly beneficial for the MCP algorithm selection task.

\begin{table*}[t]
\renewcommand{\arraystretch}{1.25}
\centering
\caption{Comparison of traditional and graph-based baselines (mean $\pm$ std; best results in \textbf{bold}; *: paired $t$-test vs.\ GAT-MLP, $p<0.05$).}
\label{tab:baselines_compare}
\setlength{\tabcolsep}{4pt}
\begin{tabular}{l c c c}
\toprule
\textbf{Model} & \textbf{Acc (\%)} & \textbf{Macro-F1} & \textbf{Weighted-F1} \\
\midrule
RF & $82.17\pm0.97^{\text{*}}$ & $0.674\pm0.053$ & $0.834\pm0.009$ \\
\midrule
GCN~\cite{gcn}           & $60.00\pm12.73^{\text{*}}$ & $0.324\pm0.136^{\text{*}}$ & $0.607\pm0.105^{\text{*}}$ \\
GCN-Global    & $85.22\pm4.46$  & $0.430\pm0.028^{\text{*}}$ & $0.797\pm0.058^{\text{*}}$ \\
GCN-GFEN      & $87.39\pm6.23$              & $0.408\pm0.109^{\text{*}}$ & $0.822\pm0.089$  \\
GIN~\cite{gin,song2023revisit}           & $53.04\pm8.22^{\text{*}}$  & $0.322\pm0.038^{\text{*}}$ & $0.589\pm0.068^{\text{*}}$ \\
GIN-Global    & $83.91\pm9.04$  & $0.452\pm0.014^{\text{*}}$ & $0.827\pm0.071^{\text{*}}$  \\
GIN-GFEN      & $79.13\pm9.55$  & $0.424\pm0.103^{\text{*}}$ & $0.787\pm0.069^{\text{*}}$ \\
SAGE~\cite{sage}          & $69.13\pm8.88^{\text{*}}$  & $0.284\pm0.080^{\text{*}}$ & $0.648\pm0.040^{\text{*}}$ \\
SAGE-Global   & $84.35\pm4.18$  & $0.424\pm0.020^{\text{*}}$ & $0.793\pm0.048^{\text{*}}$ \\
SAGE-GFEN     & $84.78\pm4.07$  & $0.350\pm0.115^{\text{*}}$ & $0.781\pm0.059^{\text{*}}$ \\
\midrule
\textbf{GAT-MLP} & $\mathbf{90.43\pm3.95}$ & $\mathbf{0.690\pm0.048}$ & $\mathbf{0.892\pm0.045}$ \\
\bottomrule
\end{tabular}
\end{table*}

Table~\ref{tab:main_results} compares our learned solver selector against four exact maximum-clique solvers executed under the prescribed time limits, the Single Best Solver (SBS), and the Virtual Best Solver (VBS). The VBS serves as an oracle selector that applies the same per-instance selection criterion used for label construction: maximizing clique size first and breaking ties by runtime. A suite of traditional ML selectors is also included as standard algorithm-selection baselines; these baselines already clearly outperform the SBS, validating the value of instance-aware selection even with simple models. Timeouts are handled via capped runtime (set to the corresponding time limit, see Section~\ref{sec:datasets}), and timeout rates are reported accordingly. Among individual solvers, CliSAT attains the highest mean clique size, 18.024, and is therefore selected as the SBS.

Our selector incurs negligible prediction overhead: predicting the solver takes $3.215\pm0.348$~seconds for the entire dataset, i.e., less than 0.01 seconds per instance on average, which is below 0.1\% of the average solving time and therefore is omitted from the reported runtimes. Across five random seeds, GAT-MLP achieves an average clique size of $20.477\pm0.226$ and an average runtime of $21.342\pm4.609$~seconds, representing a 13.6\% improvement in clique size and a $9.4\times$ speedup over the SBS. Moreover, the performance gap among solvers on the same instance can be substantial: some algorithms terminate within seconds, whereas others may run for hours or exceed the allotted time budget. Consequently, even a small number of timeouts can substantially affect the mean runtime. Our selector reduces the timeout rate to 0.79\%, a 77.4\% relative reduction compared with the 3.50\% timeout rate of the SBS. Compared with the best traditional ML model, RF, GAT-MLP achieves a 1.1\% higher mean clique size and is nearly twice as fast, demonstrating the complementary value of learned graph-topological representations beyond global statistical features. Although RF achieves a slightly lower timeout rate (0.53\% vs.\ 0.79\%), GAT-MLP offers a more favorable overall trade-off by improving both solution quality and runtime. The VBS provides an oracle upper bound on per-instance selection. GAT-MLP closes much of the gap in solution quality, achieving a mean clique size of 20.477 compared with 21.127 for VBS, while being substantially faster than all individual solvers.

For a more conservative comparison with SBS, we further include the one-time cost of computing the graph-level features used by the MLP branch, which averages 2.922 seconds per graph. After adding this overhead, the effective runtime becomes $24.264\pm4.609$ seconds per instance, still yielding an $8.27\times$ speedup over the SBS runtime of 200.636 seconds.

\begin{table*}[t]
\renewcommand{\arraystretch}{1.25}
\centering
\caption{Overall performance comparison on the maximum clique dataset. All selectors report mean~$\pm$~std over five independent runs.}
\label{tab:main_results}
\begin{tabular}{l c c c}
\toprule
\textbf{Method} & \textbf{Mean Clique $\uparrow$} & \textbf{Mean Time (s) $\downarrow$} & \textbf{Timeout Rate (\%) $\downarrow$} \\
\midrule
\multicolumn{4}{c}{\textit{Single solvers}} \\
CliSAT          & $18.024$ & $200.636$ & $3.50$ \\
LMC             & $17.394$ & $203.073$ & $4.16$ \\
MoMC            & $17.363$ & $455.981$ & $24.73$ \\
dOmega          & $15.282$ & $627.873$ & $12.69$ \\
\midrule
\multicolumn{4}{c}{\textit{Selectors}} \\
SBS (CliSAT)    & $18.024$ & $200.636$ & $3.50$ \\
Oracle (VBS)    & $21.127$ & $10.427$ & $0.00$ \\
RF              & $20.252 \pm 0.099$ & $42.264 \pm 1.199$ & $\mathbf{0.53 \pm 0.20}$ \\
KNN             & $19.969 \pm 0.179$ & $52.379 \pm 29.598$ & $1.09 \pm 0.41$ \\
DT              & $19.990 \pm 0.085$ & $42.768 \pm 10.841$ & $0.79 \pm 0.12$ \\
SVM             & $19.983 \pm 0.171$ & $47.359 \pm 14.405$ & $0.96 \pm 0.37$ \\
\textbf{GAT-MLP} 
& $\mathbf{20.477 \pm 0.226}$ 
& $\mathbf{21.342 \pm 4.609}$ 
& $0.79 \pm 0.40$ \\
\bottomrule
\end{tabular}
\end{table*}

Table~\ref{tab:family_breakdown} confirms that solver superiority is family-dependent (Fig.~\ref{fig:motivation}). The global SBS remains competitive on Random graphs but degrades elsewhere, particularly on Structured instances where the timeout rate reaches 28.95\%, and on Artificial Hard instances where it exceeds 8\%. GAT-MLP recovers performance by learning to switch to complementary solvers when local graph structure makes CliSAT brittle. On Structured graphs, whose dense community cores often favor MoMC, the selector achieves a 50.0\% larger mean clique size and a $23.8\times$ speedup, reducing the timeout rate from 28.95\% to 4.74\%. On Sparse graphs, the selector correctly identifies lightweight solvers that handle these instances almost instantly, matching the VBS in mean clique size while running $43.6\times$ faster than the SBS. Artificial Hard instances, designed to resist all solvers, see more modest but consistent gains in both mean clique size and timeout rate. Although the selector still trails the VBS on these two hard families, it closes the gap substantially on Structured graphs. These results indicate that the selector does not simply memorize solver rankings; it leverages topological signals to exploit family-specific complementarity.

\begin{table*}[t]
\renewcommand{\arraystretch}{1.18}
\centering
\caption{Performance breakdown by graph family. Global SBS refers to CliSAT. GAT-MLP results are reported as mean~$\pm$~std over five random seeds. Gain vs.\ SBS shows clique-size improvement and runtime speedup relative to the global SBS within each family.}
\label{tab:family_breakdown}
\begin{tabular}{l l c c c c}
\toprule
\textbf{Family} & \textbf{Method} 
& \textbf{Mean Clique $\uparrow$} 
& \textbf{Mean Time (s) $\downarrow$} 
& \textbf{Timeout Rate (\%) $\downarrow$}
& \textbf{Gain vs.\ SBS} \\
\midrule
\multirow{3}{*}{Random} 
& Global SBS & 14.51 & 14.48 & 0.00 & -- \\
& VBS & 14.51 & 14.02 & 0.00 & -- \\
& GAT-MLP & 14.51 $\pm$ 0.00 & 14.44 $\pm$ 0.40 & 0.00 & +0.0\%, $1.0\times$ \\
\cmidrule{2-6}
\multirow{3}{*}{Structured} 
& Global SBS & 38.47 & 2094.54 & 28.95 & -- \\
& VBS & 61.71 & 3.20 & 0.00 & -- \\
& GAT-MLP & 57.69 $\pm$ 1.23 & 88.12 $\pm$ 21.04 & 4.74 & +50.0\%, $23.8\times$ \\
\cmidrule{2-6}
\multirow{3}{*}{Sparse} 
& Global SBS & 14.49 & 11.83 & 0.59 & -- \\
& VBS & 15.41 & $<0.01$ & 0.00 & -- \\
& GAT-MLP & 15.41 $\pm$ 0.00 & 0.27 $\pm$ 0.07 & 0.00 & +6.3\%, $43.6\times$ \\
\cmidrule{2-6}
\multirow{3}{*}{Artificial Hard} 
& Global SBS & 33.49 & 201.78 & 8.11 & -- \\
& VBS & 39.54 & 109.18 & 0.00 & -- \\
& GAT-MLP & 35.64 $\pm$ 2.44 & 153.84 $\pm$ 41.93 & 4.86 & +6.4\%, $1.3\times$ \\
\bottomrule
\end{tabular}
\end{table*}

\subsection{Ablation Study}
\label{sec:ablation_study}

To quantitatively assess the incremental contribution of each component in our proposed GAT-MLP framework, we design ablation experiments to investigate two key aspects: (i) the impact of removing major architectural modules, and (ii) the performance of different fusion mechanisms.  All variants are trained under identical hyperparameters and dataset partitions, with results reported across five random seeds (mean $\pm$ std) to ensure statistical reliability.
Table~\ref{tab:ablation_results} summarizes all results.

\begin{table*}[t]
\renewcommand{\arraystretch}{1.25}
\centering
\caption{Ablation of major components and fusion strategies. All metrics are averaged over five runs (mean~$\pm$~std).}
\label{tab:ablation_results}
\begin{tabular}{l c c c c}
\toprule
\textbf{Category} & \textbf{Variant} & \textbf{Acc (\%)} & \textbf{Macro-F1} & \textbf{Weighted-F1} \\
\midrule
\multirow{3}{*}{Component}
& MLP-Only   & $83.48 \pm 8.36$ & $0.518 \pm 0.136$ & $0.832 \pm 0.072$ \\
& GAT-Only   & $56.52 \pm 11.40$ & $0.286 \pm 0.127$ & $0.604 \pm 0.093$ \\
& \textbf{GAT+MLP (full)} & $\mathbf{90.43 \pm 3.95}$ & $\mathbf{0.690 \pm 0.048}$ & $\mathbf{0.892 \pm 0.045}$ \\
\midrule
\multirow{3}{*}{Fusion}
& Weighted Fusion & $88.70 \pm 3.57$ & $0.535 \pm 0.121$ & $0.849 \pm 0.044$ \\
& Gated Fusion    & $87.83 \pm 3.64$ & $0.507 \pm 0.124$ & $0.838 \pm 0.046$ \\
& \textbf{Concatenation (ours)} & $\mathbf{90.43 \pm 3.95}$ & $\mathbf{0.690 \pm 0.048}$ & $\mathbf{0.892 \pm 0.045}$ \\
\bottomrule
\end{tabular}
\end{table*}

As presented in Table~\ref{tab:ablation_results}, excluding either the global-feature encoder (MLP-Only) or the graph-structure encoder (GAT-Only) leads to a marked performance drop, confirming their highly complementary contributions. The particularly low scores of GAT-Only further highlight that, in this task, most predictive signal resides in the global statistical descriptors rather than in raw node features and topology alone.
Among the tested fusion strategies, direct concatenation consistently outperforms both weighted and gated mechanisms, delivering the highest accuracy (90.43\%) and Macro-F1 (0.690) with the lowest variance across random seeds. We attribute this advantage to the fact that the two encoded representations are already semantically well-aligned; additional parametric fusion may therefore become redundant and, on datasets of moderate size, potentially counterproductive due to increased risk of overfitting and reduced training stability.
These findings demonstrate that, given the current dataset scale and target task, simple concatenation provides the most effective and robust approach for integrating local topological patterns and global statistical descriptors.

\subsection{Hyperparameter Analysis}
\label{sec:hyper_analysis}

We focus on two critical hyperparameters that significantly impact model performance: the learning rate ($lr$) and the dropout rate ($d$). $lr$ trades off convergence speed against optimization stability, while $d$ serves as a key regularization factor to mitigate overfitting.

We conduct a grid search over $lr \in \{10^{-4}, 5 \times 10^{-4}, 10^{-3}, 5 \times 10^{-3}, 10^{-2}\}$ and $d \in \{0.3, 0.5, 0.7\}$. For each configuration, the model is trained with five different random seeds and evaluated on the test set ensuring the reliability and reproducibility of the results.

\begin{figure*}[t]
    \centering
    \includegraphics[width=0.95\linewidth]{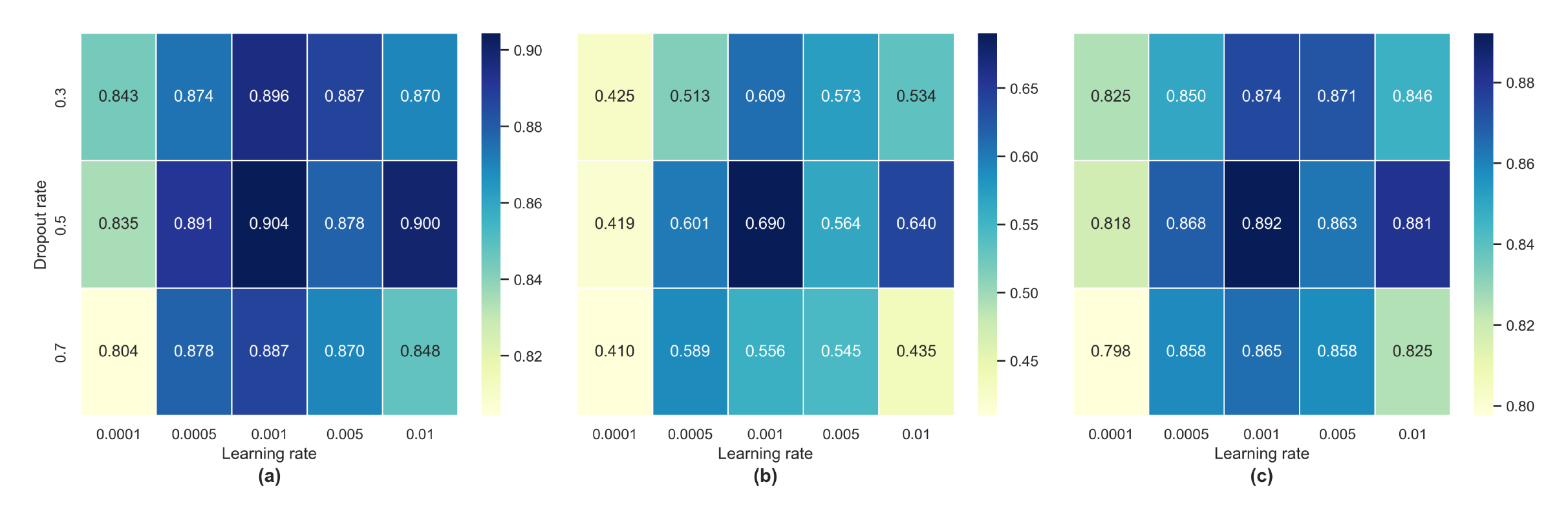}
    \caption{%
        Heatmaps of (a)~Accuracy, (b)~Macro-F1, and (c)~Weighted-F1 across learning-rate and dropout-rate combinations. 
        Darker hues denote higher metric values.
    }
    \label{fig:hparam_heatmaps}
\end{figure*}

As shown in Figure~\ref{fig:hparam_heatmaps}, the model is highly sensitive to $lr$: the test performance degrades significantly when $lr > 5 \times 10^{-3}$, while it remains consistently high for $lr \le 10^{-3}$. Moderate dropout ($d=0.5$) achieves an effective balance between fitting and regularization; $d=0.3$ offers insufficient regularization, whereas $d=0.7$ excessively hinders the model's capacity to capture task-relevant patterns.

The optimal hyperparameter configuration ($lr=10^{-3}$, $d=0.5$) corresponds to the darkest regions in the heatmaps and achieves the highest mean scores in all three metrics, demonstrating both strong predictive performance and stability with respect to random initialization.

\section{Conclusion}
\label{sec:conclusion}

In this work, we address the underexplored problem of algorithm selection for the MCP by proposing a dual-channel instance-aware framework. We construct a dataset of diverse graph instances, extract both local topological and global statistical features, and evaluate several machine learning classifiers. Building on this foundation, we design the GAT-MLP model, which integrates Graph Attention Networks with multilayer perceptrons to jointly capture structural and statistical cues. Experimental results reveal that our method achieves significant performance gains over traditional baselines, alongside three representative GNN architectures (GCN, GIN, SAGE) and their enhanced variants, in the task of MCP algorithm selection. Despite these promising results, the study still has two practical limitations: the dataset size is modest and the set of MCP solvers considered is limited to four representative algorithms. Future work will focus on expanding the dataset with more diverse graph instances, incorporating weighted and approximate MCP solvers, and exploring ways to transfer the proposed framework to related combinatorial optimization problems.

\bibliographystyle{IEEEtran}
\bibliography{references}

\end{document}